# Ontologies-based Architecture for Sociocultural Knowledge Co-Construction Systems

**Guidedi Kaladzavi**, University of Maroua, Maroua, Cameroon kaladzavi@univ-maroua.cm

**Papa Fary Diallo †**, University of Gaston Berger, Saint-Louis, Senegal

**Cedric Bere**, University of Ouaga 1, Ouagadougou, Burkina Fasso, cedric.bere@gmail.com

**Olivier Corby**, Inria, UCA, Sophia Antipolis, France, olivier.corby@inria.fr

**Isabelle Mirbel**, UCA, Sophia Antipolis, France, isabelle.mirbel@unice.fr

**Moussa Lo**, University Gaston Berger, Senegal, moussa.lo@ugb.edu.sn

**Kolyang,** University of Maroua, Maroua Cameroon, kolyang@univ-maroua.cm

## Abstract

*Considering the evolution of the semantic wiki engine based platforms, two main approaches could be distinguished: Ontologies for Wikis (OfW) and Wikis for Ontologies (WfO). OfW vision requires existing ontologies to be imported. Most of them use the RDF-based (Resource Description Framework) systems in conjunction with the standard SQL (Structured Query Language) database to manage and query semantic data. But, relational database is not an ideal type of storage for semantic data. A more natural data model for SMW (Semantic MediaWiki) is RDF, a data format that organizes information in graphs rather than in fixed database tables. This paper presents an ontology based architecture, which aims to implement this idea. The architecture mainly includes three layered functional architectures: Web User Interface Layer, Semantic Layer and Persistence Layer.*

## Introduction

This research study is set in an African context, where the main problem is an economic, social development and the means to achieve it. Indeed, after the failure of several development models in the recent decades, theoretical research seems to be turning to the development knowledge-based approaches (UNESCO, 2014). The place of knowledge, science and technology in the current dynamics of growth gives rise to intensify the reflection within the economic field. In particular, many authors point out that we entered in the new phase of knowledge-based development following a phase of the physical capital accumulation.

To promote the indigenous knowledge, some media could be used : a permanent (re)education, the radio broadcasting, and off course the Internet, which seems to be the best media (UNESCO, 2014). It reduces (instantaneously) distances between civilizations. Thus, it is an opportunity to disseminate the local knowledge on a large scale. But, this is not sufficient to make the Internet the ultimate solution to the African culture vulgarization. It does not create anything itself. It is the

African responsibility to build the content of the "empty shell" that is the Internet and make the rational use and enjoy the opportunities it offers.

To get there, new computational technologies (semantic technologies) are needed to manage these large repositories of sociocultural data and to discover some useful patterns and knowledge from them. Semantic Web vision proposed by Tim Berners-Lee, revolutionized the Web architecture. The Web architecture switched from the documentary graph to the published and interconnected databases with capabilities to "understand" their semantics and reason on them. Technically, are introduced into the Web architecture stack, the RDF data model and the URI standard for modeling and identifying resources on the Web. As result, the Web is spreading in the World and the World is spreading in the Web with issues such as cultural "digital divide". Indeed, a cultural void in the Web of Data is the lack of that culture at the applicative level (e-tourism, etc.) of the Web of Data.

In this paper, we present the architecture of a sociocultural knowledge co-construction platform, developed to enable some communities to share and co-build their cultural heritage. Thus, the architecture relies essentially on the USCO (Upper-Level Sociocultural Ontology) (Diallo et al., 2014) ontology, which allows communities to share and co-construct their socio-cultural knowledge. Since the USCO ontology is aligned with the vocabularies of Schema.org and DBpedia, the latter are also integrated and used in the platform. To allow temporal annotation of resources, the HuTO (Human Temporal Ontology) ontology is used (Diallo et al., 2015). Ontoshare (Kaladzavi et al., 2015) models how contributors could co-build the content of the KMS (Knowledge Management System) circumscribed by USCO. The functionalities of the platform could also allow members of the African communities (Senegalese in particular) to co-elaborate knowledge, to exchange and compare their points of view in the process of co-construction of knowledge.

The rest of the paper is schemed as follows. In Section II we present the definition of some core concepts. Section III, entitled related work reviews some existing studies devoted to the Sociocultural Knowledge Management. Then, Section IV depicts the proposed architecture. The Architecture presentation consists of presenting its main layers. In Section V, we point out the performance evaluation feedback. The paper ends with a conclusion and future trends in Section V.

## Conceptual Framework

We present in this section the definition of some core concepts, which could facilitate the understanding of the paper.

### Culture

Culture consists of that complex whole, which includes knowledge, belief, art, morals, law, custom and any other capabilities and habits acquired by man as a member of a society (Tylor, 2005). There are many definitions regarding the domain (Philosophy, Sociology, etc.). In sociology, culture is defined more narrowly as "what is common to a group of individuals" that is to say what is learned, transmitted, produced and created. For UNESCO (United Nations for Educational, Scientific and Cultural Organization), culture may now be regarded as the set of distinctive spiritual, material, intellectual and emotional features that characterize a society or social group. It encompasses the arts, literature and science, lifestyles, fundamental rights of the human being, value systems, traditions and beliefs (UNESCO, 1982).

## Sociocultural Knowledge

The sociocultural qualifier links the belief, art, morals, law, custom and any other capabilities and habits to a society. In other words, the qualifier links to a group of people (European, African, Asian, etc.) who shares the same experiences. Sociocultural knowledge describes information about society and culture. It is related to, or involves a combination of social and cultural factors. It acquires its richness in an environment where several people share, a space (continent, country, region, state, locality) in which there are interactions between cultures.

Being interested by the sociocultural knowledge in the sense of its valorization and transmission to the future generations, is to focus on the heritage aspects of a society, a people or a country. Heritage territories are plural and fuzzy (Benhamou, & Thesmar, 2011). Since 1986, A. Chastel already mentioned that the concept of heritage is global, vague and pervasive (Benhamou, & Thesmar, 2011). Some experts but also the public define what is intended to be preserved. It is therefore subject to possible redefinitions. Then the heritage perimeter is drawn by UNESCO, an international authority, which establishes criteria and a qualified list of heritages. According to UNESCO, heritage means what a country intends to preserve for the future generations. It includes both a report on the history and the future, because of its continuities (with the benefit of hindsight that brings almost naturally new monuments in the heritage field) and discontinuities (with the introduction of new objects and concepts that expand the field of heritage, gardens, landscapes, industrial sites, public infrastructures (schools, airports, hospitals, etc.), various memorial sites, etc.)). But also intangible heritage such as the traditional festivities, sport activities, religious activities, etc., that contribute to the development of what could be called a "national romance".

## Ontologies

The foundational definition of ontology is proposed by Gruber (Gruber, 1993; 2005): An ontology is "an explicit specification of a conceptualization". The exact meaning depends on the understanding of the terms "specification" and "conceptualization". According to (Genesereth & Nelson, 1987), conceptualization is a "set of objects, concepts, and other entities that are presumed to exist in some domains of interest and the relationships that hold them". In Gruber's definition, it is not clear that specification depends on the logical view of ontologists. That is why Guarino and Giaretta introduced the logical theory instead of mere specification (Camara, 2013). Afterward, Borst enriches the previous definition by adding consensual facts related to knowledge modeling discipline characteristics such as sharing and reuse (Camara, 2013). For him, "Ontologies are defined as a formal specification of a shared conceptualization". Finally, Studer and collaborators merge the existing definitions (Studer et al., 1998). For them, "An ontology is a formal and explicit specification of a shared conceptualization". They underline the necessity of formal, explicit and shared paradigms. Even if it is the merging of the existing definitions, it seems consensual. It is more cited in recent years, demonstrating its compliance with the expectations of Knowledge Base systems designers (Camara, 2013). The explicitly, formality and shareability knowledge features in an ontology are carried out by five elements: concepts, relations, functions, axioms, and instances (Gomez-Pérez, 1999).

## Co-construction

The term co-construction is an innovative term, which burst into everyday language recently: used in print media once a year before 2003, once a month in 2005, it appears almost daily in 2013

(Akrich, 2013). Typically, it is used to enhance the involvement of the plurality of actors in the development and implementation of a project or an action. In the academic literature, the term has undergone a parallel evolution to that, which is observed in the press. In Educational Sciences, it denotes the desire to get out of a vertical transmission of knowledge by actively and collaboratively involving pupils or students in the production of learning content through ICT. In sociology or political science, it means the existence of the plurality of actors involved into the production of a policy, project, a technical device or knowledge. In the case of Human-Computer-Interaction (HCI), it implements how users interact among themselves to collaboratively achieve a virtual activity through software interface. It is the fundamental idea behind the Web 2.0 vision, which democratizes the knowledge production on the Web. By Web 2.0, Internet users are no longer just consumers but also authors of information. Thus, the social Web enables Internet users to share knowledge, that is to externalize their expertise about something or to learn (internalize) what other Internet users have published.

## Related Work

Since the Web 2.0, the flows have been reversed: the user is no longer passive (reader) but active (author). The transition from Web 1.0 to Web 2.0 done in 2004 was a decisive articulation to social knowledge management systems (Social media). Social media includes tools and applications that allow interaction between Internet users. Within that galaxy of social media, there are several "planets". Out of them, there are texts publishing tools (wiki, blogs, etc.), exchange and sharing tools (YouTube for videos, Slideshare for sharing presentations, etc.), tools for discussion (Skype, Messenger, etc.) and the networking implementation tools (Twitter, Facebook, Myspace, etc.), supply is abundant! Unfortunately, most of them are not local-knowledge oriented. Some cases exist, such as Wikipedia and Afripedia. Afripedia is a project devoted to develop Wikipedia in the African context. It is the main project launched in 2012 by the "Agence Universitaire de la Francophonie (AUF)", the French Institute (IF) and the French Wikimedia Association, which aims to africanize the Wikipedia encyclopedia by integrating the African knowledge in the platform, even where the Internet access is not possible (that means in the offline way). In fact, before the Afripedia project, Wikimedia association was worried about the few contributions from Africa and relative to Africa. In geographic sections for example, one could even see that a mountain somewhere in African locality was described as a hill. In addition, the platform does not use semantic technologies, which means that it is not ontology based (kaladzavi et al., 2015).

The management of knowledge is increasingly being recognized as a key element in extracting its value. Knowledge Management is a discipline that provides strategy, process and technology to share and leverage information and expertise that will increase our level of understanding to (more) effectively solve problems, and make decisions (Satyadas et al., 2001). Knowledge Management role can be viewed as turning data into information and then forming information into knowledge. It is largely regarded as a cyclic process involving various activities (Nonaka, 1999). The process can be subdivided, for example into creating internal knowledge, acquiring external knowledge, storing knowledge as well as updating the knowledge and sharing knowledge internally and externally (Alavi & Leidner, 2001). The Knowledge Management System (KMS) architecture is a fundamental issue in the area of Knowledge Management that must be well resolved in order to deliver competitive services to the users as well as the organization. (Meso & Smith, 2002) proposed a KMS architecture that processes a combination of all the aspects stated below as well as other components, which is able to perform according to the requirement of the organization. These components consist of technology functions and knowledge itself. Different factors and constraints influence the development of an architecture: functional requirements, quality considerations, experience and technical aspects (Jacobson et al., 1999). In the context of Web of Data, the architecture is primarily influenced by functional requirements, that means the services provided, and the quality considerations such as scalability, performance, reusability and semantic interoperability. These constraints are data models based-on. Thus, the semantic layer compound by RDF, RDFS, RIF /SWRL, SPARQL and OWL technologies are key components in the architecture design.

## Architecture

We proposed in our previous work how we can create, acquire, store, as well as update and share the sociocultural knowledge internally and externally by modeling some sociocultural ontologies:

- **USCO** : USCO ontology is used to enable communities to share and co-build their sociocultural heritage. This is done through the descriptions made of the individuals of sociocultural events organized by these communities, the descriptions of the available resources to these communities but also the descriptions made about these communities themselves.

- **HuTO** : HuTO models the deictic dates, which are dates that form a specific relation with the time of the discourse. In our modeling, deictic times are sub-concepts of the Date concept. Indeed, one cannot know to what refers a deictic date without knowing the temporal position of the discourse. The modeling choice, is to model the deictic time by associating the properties of the effective date associated with the deictic time.

- **Ontoshare** : Ontoshare is a virtual activity ontology in the case of the sociocultural knowledge sharing. It designs how Internet users could share and co-construct the content of a sociocultural Knowledge Management System (KMS).

The developed architecture is based on three entities: 1) Semantic MediaWiki (SMW), which is the base of the platform, 2) Virtuoso, which is the triple store (RDF database) and Exhibit, which is the data visualization tool. Figure 1 shows the different layers of the architecture. This architecture consists of three layers: A Human-Machine-Interface (HMI) supported by SMW and Exhibit. This layer allows users to participate in the co-construction of knowledge. It also makes it possible to present the data in a user-friendly way (Exhibit). The semantic layer makes it possible to have a query interface and a rules engine. The persistence layer consists of a relational database and a triple store (semantic database).

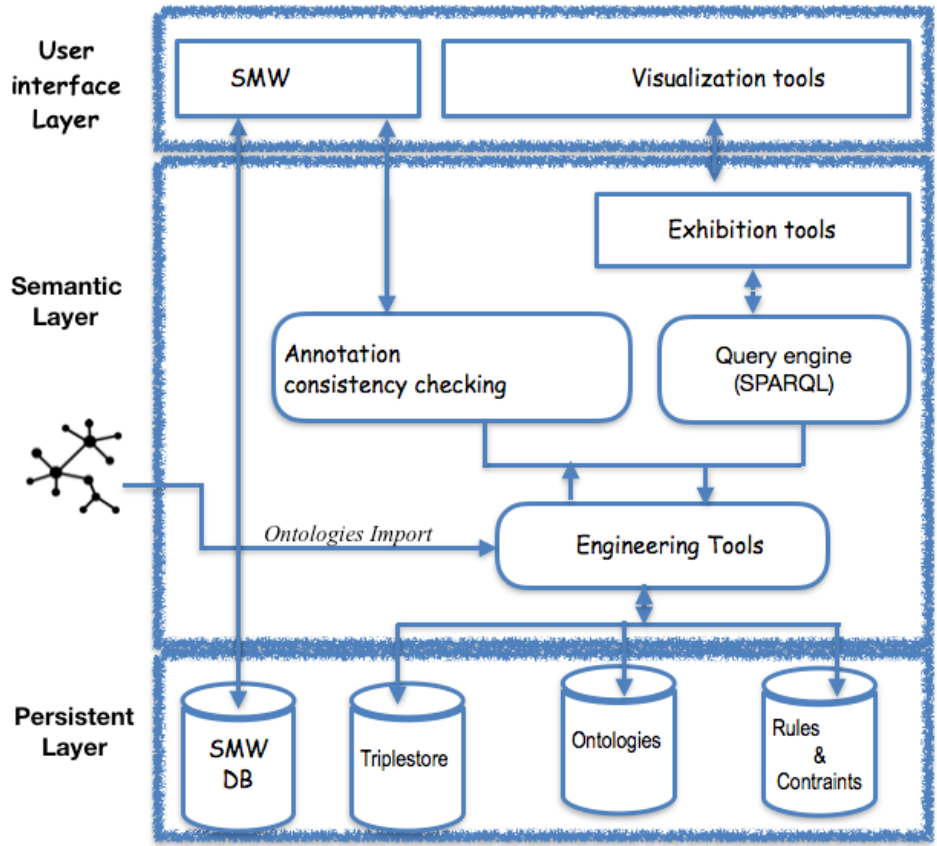

**Figure 1:** System Architecture

## Web User Interface Layer

The HCI layer is composed of the implemented tools, so that a human can control and communicate with the platform. The first tool is SMW, an extension of MediaWiki (MW). The second tool is Exhibit, which belongs to the SIMILE (Semantic Interoperability of Metadata and Information in unLike Environments) project (MIT, 2006).

### *Semantic MediaWiki (SMW)*

Semantic MediaWiki (SMW) is an extension of MediaWiki, the Wikipedia-based wiki engine (semantic-mediawiki.org, 2017). Unlike traditional wikis that can only contain texts that can be not 'understood' or evaluated by software agents, SMW adds semantic annotations to the Wiki pages using an extension of the MediaWiki script language. Thus, SMW enables MediaWiki to provide functions such as a collaborative knowledge base. MW has a scripting language to describe the content of the Wiki pages. This language was extended by SMW with the following three sets of semantic annotations: Classes, Properties, Axioms and Instances. Thus, the collection of semantic data in SMW is made by allowing users to add annotations on the articles pages through special tags. Each item corresponds exactly to an ontological element, that is to the following RDF types: owl:Thing, rdfs:Class, owl:ObjectProperty, owl:DatatypeProperty and owl:AnnotationProperty.

In particular, an article is an instance of owl:NamedIndividual, categories become classes and properties, depending on their type in the Wiki (See Table 1).

**Table 1:** Mappings between SMW and OWL

| SMW | Syntax | OWL |
|---|---|---|
| Relation | [ [ Capital::Senegal] ] | owl:ObjectProperty |
| Attribute | [ [ Population::1056009] ] | owl:DatatypeProperty |
| Category (in article) | [ [ Category:City] ] | owl:NameIndividual |
| Category (in a namespace) | [ [ Category::Locality] ] | rdfs:subClassOf |

### *Exhibit*

Exhibit is developed during the SIMILE project, a collaboration between MIT (Massachusetts Institute of Technology), the Artificial Intelligence Lab (CSAIL) and the World Wide Web Consortium (W3C). The goal of the SIMILE project is to develop tools to increase the interoperability of disparate digital collections and develop robust open source tools that allow users to access, manage, visualize and reuse schemas/vocabularies/ontologies and metadata (MIT, 2006). Exhibit is part of these tools and allows to create Web pages with a support of sorting, filtering and rich visualizations by using HTML, CSS and JavaScript. The Exhibit user interface consists of three main component types: Collection, Widget and Coder.

### Semantic Layer

The main role of the semantic layer is to link the Knowledge Base: 1) to Exhibit for better visualization, 2) to the external resources to feed the Knowledge Base, and 3) to the defined rules to deduce implicit relations in the Knowledge Base. All these actions are done through the Jena framework. It provides an API for retrieving and writing data in RDF graphs. The graphs are represented as an abstract model. A template can come from data, files, databases, URIs, or a combination of them. A model can also be queried by a SPARQL endpoint. Thus, to allow access to the data stored in Virtuoso, the triple store provides a Jena RDF Data Provider, which is a service for the Jena framework and allows Semantic Web applications to directly query a Virtuoso Knowledge Base. Query Engine (See Figure 1) is a SPARQL endpoint interface provided by Virtuoso. This interface allows users to directly access Virtuoso data. However, thanks to the Jena framework, Web applications are developed to access the data of the platform. Rules Engine is a Java EE component that is developed and that executes these rules. However, we used in these rules the INSERT clauses of SPARQL in place of the CONSTRUCT clauses.

## Persistence Layer

Typically, SMW stores data in a Relational Database. However, for RDF data, the best option is to use a data format that organizes information in the form of a triples rather than in relational database tables. Indeed, the use of a triple store has several advantages :

1. **Better performance for information retrieval:** RDF databases are configured to respond to SPARQL queries, which is a W3C recommendation. SPARQL offers better performance in information retrieval than the use of SQL queries in a relational database. Indeed, SPARQL queries make it possible to benefit from the semantics of the RDF data

2. **An additional interface:** with a SPARQL endpoint (Web query interface), it is possible to make SPARQL queries on the data without resorting to the interface of SMW. This allows other applications to access the data. A reasoning functionality: semantic Web languages such as RDFS and OWL offer powerful modeling features such as subsumption relations between the concepts or characteristics of certain properties as transitivity. Thus, an RDF database can infer on these characteristics to get answers. The reuse of ontologies and the integration of data: it is possible to add the external data in the RDF database and to use SMW to update them. Thus, the RDF database can be seen as a platform for data integration and ontologies reuse.

3. **Physical separation of resources:** Separating the data used by MW from SMW provides a way to distribute tasks across multiple servers. In particular, complex queries, which consume a large amount of computational power, may thus not affect the basic operation of the Wiki. Thus, SMW can operate with five different triple stores. Table 2 compares these triple stores. A graph database represents the data in graph structures composed of nodes and edges. This allows easy processing of the data by a calculation of the specific properties of the graph, such as the number of steps (nodes) needed to go from one node to another. A triple store implements a graph model that interprets the predicate as the label of a link between subject and object.

Among the triple stores (see Table 2), Virtuoso has been used. It is a hybrid database and a middleware that supports relational data, graphical and RDF data, web-server applications, and so on. By integrating Virtuoso with SMW, only the semantic data is migrated into Virtuoso. As shown in Figure 1, all textual data remains in the MySQL database that is used by MediaWiki, and all annotations made in the Wiki are stored in Virtuoso. For the better scaling, the database is divided into three warehouses (graphs): 1) a warehouse where the semantic data are stored (annotations from the Wiki), 2) a warehouse where the USCO ontology is stored, and 3) a warehouse , where the HuTO ontology is stored. Since SMW allows importing vocabularies, the vocabularies of HuTO and USCO are imported into the platform. This allows users to annotate the resources of the Wiki using the vocabularies of HuTO, USCO and Ontoshare.

|  | **Blazegraph** | **Jena TDB** | **Sesame** | **Virtuoso** | **4store** |
|---|---|---|---|---|---|
| **Supported Model** | Graphic and RDF store | RDF stores | RDF stores | Graphic And RDF stores | RDF stores |
| **Storable triple stores** | 12,7B | 1,7B | 70 M | 15,4B | 15B |
| **RDF store classification** | 15 | 3 | 4 | 2 | 8 |

**Table 2:** Comparison of the triple stores used by SMW (M stands for million and B stands for billion.) For example, DBpedia has less than 440M triples.

## Performance Evaluation

### Semantic Interoperability

The proposed architecture enables the use of external resources to feed the Knowledge Base. These resources could be imported from the Linked Open Data, that is to say open data for which, a SPARQL endpoint interface exists. The English version of DBpedia was used to extract data related to Senegalese case study. Indeed, DBpedia proposes a Knowledge Base that gathers several themes and data specific to countries. Thus, the alignment between the DBpedia concepts such as DBpedia:PopulatedPlace and usco:Locality enables to extract all Senegalese cities from the Knowledge Base of DBpedia thanks to the query below:

```
PREFIX ontology: <http://dbpedia.org/ontology/>
PREFIX resource: <http://dbpedia.org/resource/>
SELECT * WHERE {
  ?l ontology:type resource:Regions_of_Senegal .
  ?o ontology:isPartOf ?l
}
```

**Query 1** : Request to have the regions of Senegal and their administrative subdivisions which are in the Knowledge Base of DBpedia.

## Rules

HuTO provides a conceptual model in RDFS for modeling temporal expressions and annotating RDF resources. However, many temporal relationships are implicitly expressed in occurrences of events (relative dating). Answers to many time-oriented questions are not necessarily explicitly represented but can be deduced. To do this, it proposed a set of rules to standardize the representation of temporal data and also rules of inference. Since HuTO is an ontology in RDFS, some rules have been proposed, expressed as CONSTRUCT queries in SPARQL, with the aim of deducing and explaining the maximum temporal information to enable the reasoning capabilities. The temporal information can be expressed in different ways. For example, a date (month) can be represented either by using the calendar representation (See Query 1). Also, some rules have been created to standardize these types of writings. Therefore, whatever the writing used mode, possible representations will be added to the data graph.

```
PREFIX huto: <http://ns.inria.fr/huto/>
CONSTRUCT {
    ?x huto:number ?m ; huto:numberOfDay ?d ; huto:even ?e
}
WHERE {
    ?x rdf:type ?o
    ?o rdfs:subClassOf huto:Month ;
       huto:number ?m ;
       huto:even ?e .
    OPTIONAL { ?x rdf:type/huto:numberOfDay ?d }
}
```

**Query 2** : Example of month normalization rules.

## Inferences

However, since RDFS does not implement some basic inferences such as *transitivity* or *reflexivity*, we created some inference rules for this purpose. Thus, we defined inference rules for the transitivity of before/after properties. Similarly, if a relation (after or before) is expressed between two events (respectively intervals), it is necessary to propagate this relation between the intervals (respectively resources) concerned. For this, we proposed propagation rules. In total, 26 rules of implications and inferences were defined for properties before and after. These rules are defined for transitivity, inverse and propagation.

```
PREFIX huto: <http://ns.inria.fr/huto/>
CONSTRUCT { ?x huto:before ?y }
WHERE {
    ?s huto:before ?o .
    ?x a huto:TemporalAnnotation ;
       huto:hasTemporalExp ?s .
    ?y a huto:TemporalAnnotation ;
       huto:hasTemporalExp ?o .
    FILTER NOT EXISTS { ?x huto:before ?y }
}
```

**Query 3** : Rule for propagating the "before" property between resources.

## Semantic queries

We distinguish two types of queries: 1) A resource-type request: determines the period of occurrence of the given resource. 2) Typical queries on temporal elements: they determine resources relating to a given period of occurrence or relative to a temporally annotated resource. The SPARQL query (See Query 4) is an example of the request written from the resource-type request in which, the retrieved resource was specified. It allows to determine the temporality of the resource data:Gamou.

```
DESCRIBE ?x
WHERE {
    { ?x huto:uri ?resource } UNION
    { ?x huto:triple/(rdf:subject|rdf:object) ?resource } UNION
    { ?x huto:graph ?g .
      graph ?g {
          { ?resource ?p ?o } UNION
          { ?s ?p ?resource }
      }
    }
    FILTER NOT EXISTS { ?j ?k ?x }
}
VALUES ?resource { data:Gamou }
```

**Query 4** : Determining the temporality of the resource data:gamou

## Conclusion and Perspectives

In this paper, we presented an ontology-based architecture for sociocultural Knowledge Management Systems. The platform can be considered as a "collective memory" that enables users to share and co-build knowledge. The platform could help to capture the holistic view of the local changes while considering culture and historicity in the context of a country. The proposed architecture includes three layered functional architecture: Web user interface layer, semantic layer and persistent layer. We illustrated how Internet users can internalize or externalize knowledge in the ethical way in the Senegalese context. In addition, we evaluated the architecture performance regarding semantic interoperability, rules, inferences features. The success of the system depends on some ICT features. That is why we would like to point out that some ICT access features such as « divide by access », « divide by decision » in African Countries must be improved.

Collaborative knowledge base systems face the speculations problems. Yet, the credibility of contributors is increasingly sought after by fact-checking algorithms. As result, the future work will be focused on how to take into account the fact-checking module in our architecture.

## Acknowledgement

In memory of Papa Fary Diallo, who brutally left us while we were discussing on this paper. The authors would like to thank the CEA-MITIC (Centre d'Excellence Africain en Mathématiques, Informatique et TIC ), the University of Maria (UMa) and LANI (Laboratoire d'Analyse Numérique et d'Informatique ) for the financial support.